\documentclass[10pt,twocolumn,letterpaper]{article}
\usepackage{newtxtext,newtxmath}
\usepackage{tgtermes}  
\usepackage{amsmath}
\usepackage{graphicx}
\usepackage{hyperref}
\usepackage{calc} 
\usepackage{geometry}
\usepackage[numbers,square]{natbib}
\usepackage[utf8]{inputenc}
\usepackage[T1]{fontenc}
\usepackage{titlesec}
\usepackage{calc} 
\usepackage{times}
\usepackage{setspace}
\usepackage{indentfirst}
\usepackage{etoolbox}
\usepackage{enumerate} 
\usepackage{enumitem} 
\usepackage{comment} 
\usepackage{amsmath}   
\usepackage[font=small]{caption}         
\usepackage{algorithm}
\usepackage{algpseudocode}
\usepackage{xparse}
\usepackage{float}  
\usepackage{dblfloatfix}
\usepackage{booktabs}  
\usepackage{multirow}



\usepackage{titlesec}
\titlespacing{\section}{0pt}{1ex}{0.5ex}
\titlespacing{\subsection}{0pt}{1ex}{0.5ex}

\titlespacing{\subsubsection}{0pt}{0.5ex}{0.3ex}

\AtBeginEnvironment{algorithmic}{\fontfamily{ptm}\selectfont\small}
\renewcommand{\texttt}[1]{{\fontencoding{T1}\fontfamily{qtm}\selectfont #1}}
\NewDocumentCommand{\longstate}{O{1.5em} m m}{%
  \State \hspace*{#2}%
  \parbox[t]{\linewidth-#2}{\hangindent=#1 \hangafter=1 #3}%
}

\makeatletter
\patchcmd{\@afterheading}{\indent}{\noindent}{}{}
\makeatother

\newlength{\charwidth}
\newlength{\colsepvalue}
\newcommand{\authorName}[1]{%
  \fontsize{9}{10.8}\selectfont #1\par\vspace{2pt}%
}
\newcommand{\affiliationEmail}[1]{%
  \fontsize{9}{10.8}\selectfont #1\par\vspace{2pt}%
}

\author{
  \centering
  \begin{minipage}[t]{0.5\linewidth - 0.5\colsepvalue}
    \centering  
    \authorName{Anlong Zhang}
    \affiliationEmail{Institute of Advanced Technology\\ 
    University of Science and Technology of China (USTC)\\
    Hefei 230026, China\\
    \texttt{zhanganlong@mail.ustc.edu.cn}}
  \end{minipage}
  \hspace*{\colsepvalue}
  \begin{minipage}[t]{0.5\linewidth - 0.5\colsepvalue}
    \centering  
    \authorName{Jianmin Ji\thanks{Corresponding author: \texttt{jianmin@ustc.edu.cn}}}
    \affiliationEmail{School of Computer Science and Technology\\
    University of Science and Technology of China\\
    Hefei 230026, China}
  \end{minipage}
}

\renewenvironment{abstract}{%
  \noindent\fontsize{9}{10.8}\selectfont%
  \setlength{\parindent}{0.48cm}%
  \setlength{\parskip}{10pt}%
  \textbf{abstract}—\ignorespaces%
}{\par\vspace{10pt}}

\newcommand{\keywords}[1]{%
  \par\vspace{6pt}%
  \noindent\fontsize{9}{10.8}\selectfont%
  \setlength{\parindent}{0.48cm}%
  \textbf{keywords}—#1\par%
}

\selectfont 

\title{Research on Navigation Methods Based on LLMs}
\date{}

\begin{document}
\maketitle

\begin{abstract}
In recent years, the field of indoor navigation has witnessed groundbreaking advancements through the integration of Large Language Models (LLMs). Traditional navigation approaches relying on pre-built maps or reinforcement learning exhibit limitations such as poor generalization and limited adaptability to dynamic environments. In contrast, LLMs offer a novel paradigm for complex indoor navigation tasks by leveraging their exceptional semantic comprehension, reasoning capabilities, and zero-shot generalization properties. We propose an LLM-based navigation framework that leverages function calling capabilities, positioning the LLM as the central controller. Our methodology involves modular decomposition of conventional navigation functions into reusable LLM tools with expandable configurations. This is complemented by a systematically designed, transferable system prompt template and interaction workflow that can be easily adapted across different implementations. Experimental validation in PyBullet simulation environments across diverse scenarios demonstrates the substantial potential and effectiveness of our approach, particularly in achieving context-aware navigation through dynamic tool composition.
\end{abstract}

\keywords{Large Language Models (LLMs); Navigation; Function Calling; Dynamic Tool Composition}

\section{Introduction}
    Introduction
Navigation constitutes a fundamental challenge in robotic tasks, requiring intelligent agents to navigate toward specified targets in unknown environments. Traditional navigation approaches, while capable of accomplishing numerous tasks, present limitations in either requiring extensive training data or demonstrating insufficient generalization capabilities and semantic comprehension. The emergence of large language models (LLMs) has revolutionized autonomous planning through their exceptional capacity to comprehend and reason about complex, context-rich scenarios \cite{ref1}. Recent advancements such as NAVGPT and NAVGPT2 convert visual scene semantics into LLM-compatible input prompts, enabling visual-language navigation through LLMs' commonsense knowledge and reasoning capabilities. However, current LLM-based implementations like VLTNet with Tree-of-Thought networks for language-driven zero-shot object navigation (L-ZSON) ~\cite{ref8}, UniGoal \cite{ref3} with unified graph representations for general zero-shot navigation, and MapNav's end-to-end visual-language navigation using annotated semantic maps (ASM) as historical frame replacements, while leveraging LLM capabilities, fail to fully exploit robotic systems' inherent potential. These approaches exhibit limited adaptability to diverse environmental conditions and do not enhance robots' intrinsic generalization capabilities.

Our proposed methodology addresses these limitations through two key innovations: First, we decompose robotic functionalities into modular tools that LLMs can strategically invoke for task execution. Second, we implement a modified chain-of-thought (CoT) protocol in system prompts to ensure reliable tool invocation. A three-level safeguard mechanism guarantees operational robustness: 1) Comprehensive descriptions in tool definitions with error feedback integration during execution, 2) CoT-structured system prompts minimizing invocation errors through constrained reasoning pathways, and 3) LLMs' inherent self-correction capabilities providing final error recovery. This dual approach not only enhances environmental adaptability through robotic capability modularization but also maintains compatibility with future hardware/software advancements, while synergistically combining with LLMs' expanding potential.

This work makes four key advances: (1) We pioneer a function-calling paradigm that systematically decomposes navigation tasks into modular operations, directly harnessing LLMs' commonsense reasoning for execution—a departure from existing end-to-end approaches; (2) Our modular architecture introduces unprecedented environmental adaptability, enabling flexible customization through function modification/extension without model retraining, thus addressing cross-scene compatibility limitations; (3) The proposed LLM agent core establishes a generalizable framework for embodied intelligence applications beyond navigation, demonstrating transfer potential to diverse robotic tasks; (4) We formalize a reusable system template with standardized prompts and interaction protocols, ensuring reproducibility while maintaining customization capacity for specific deployment scenarios. These innovations collectively advance the integration of linguistic intelligence with robotic systems through structural flexibility and methodological rigor.

\section{Related work}
    The remarkable advancements in Large Language Model (LLM) training \cite{ref4,ref9} have catalyzed a notable trend of integrating LLMs into embodied robotic tasks, as exemplified by systems like SayCan \cite{ref20} and PaLM-E \cite{ref13}. This paradigm shift stems from two fundamental advantages of language models: the scalability of training data and model parameters. First, natural language processing breakthroughs enable the acquisition of cross-domain knowledge through massive textual corpora. Second, the emergent capabilities derived from scaling model architectures with virtually unlimited linguistic data \cite{ref15} significantly enhance reasoning proficiency across diverse domains. These properties position LLMs as promising foundations for developing general-purpose embodied intelligent agents.

Traditional indoor navigation systems, while achieving basic localization in controlled environments, suffer from critical limitations in generalization capability and semantic comprehension. Regarding generalization: 1) Current methods rely heavily on pre-deployed infrastructure (Wi-Fi fingerprint databases, Bluetooth beacons) and static environmental assumptions, resulting in poor cross-scene adaptability. Radio signal propagation models (e.g., RSSI attenuation) exhibit significant variance due to architectural configurations and electromagnetic interference, while dynamic perturbations (human flow, temporary obstacles) exacerbate inertial navigation drift and wireless positioning fluctuations without adaptive calibration mechanisms \cite{ref10}. In terms of semantic understanding: Conventional approaches focus on geometric path planning while neglecting environment-user context modeling, manifesting as: 1) Inability to interpret implicit requirements in natural language instructions (e.g., ``find a less crowded charging zone''), limited to predefined keyword responses; 2) Disjointed semantic reasoning from real-time decision-making due to inadequate multimodal data fusion (visual cues, voice interactions); 3) Path planning oblivious to knowledge-driven constraints (e.g., ``fire lane restrictions'', ``meeting room occupancy''), potentially violating semantic rules \cite{ref6}. These limitations highlight the necessity for cross-modal semantic modeling (vision-language joint representation learning) and adaptive frameworks (meta-learning enhanced lightweight migration) to enable environment-user synergy in intelligent navigation systems.

While LLM applications in robotics remain nascent \cite{ref4,ref7}, preliminary explorations demonstrate their potential in navigation tasks. Shah et al. \cite{ref11} leveraged GPT-3 for landmark identification, while Huang et al. \cite{ref12} focused on code generation through LLMs. Zhou et al. \cite{ref14} employed LLM-derived commonsense knowledge of object relationships for zero-shot object navigation (ZSON) \cite{ref17,ref18}, an approach paralleled by Jacky Liang et al. in programming robotic policies via natural language instructions. However, our 
work diverges by proposing a functional decomposition strategy that directly invokes LLM-powered reasoning through modular function calls. The closest precedent, NAVGPT, converts visual scene semantics into LLM prompts for vision-language navigation but suffers from critical limitations: 1) Excessive history length in LLM interactions that degrades processing efficacy; 2) Insufficient extensibility and environmental compatibility.    

\section{Method}
\begin{figure}[ht]
  \centering
  \includegraphics[width=\linewidth]{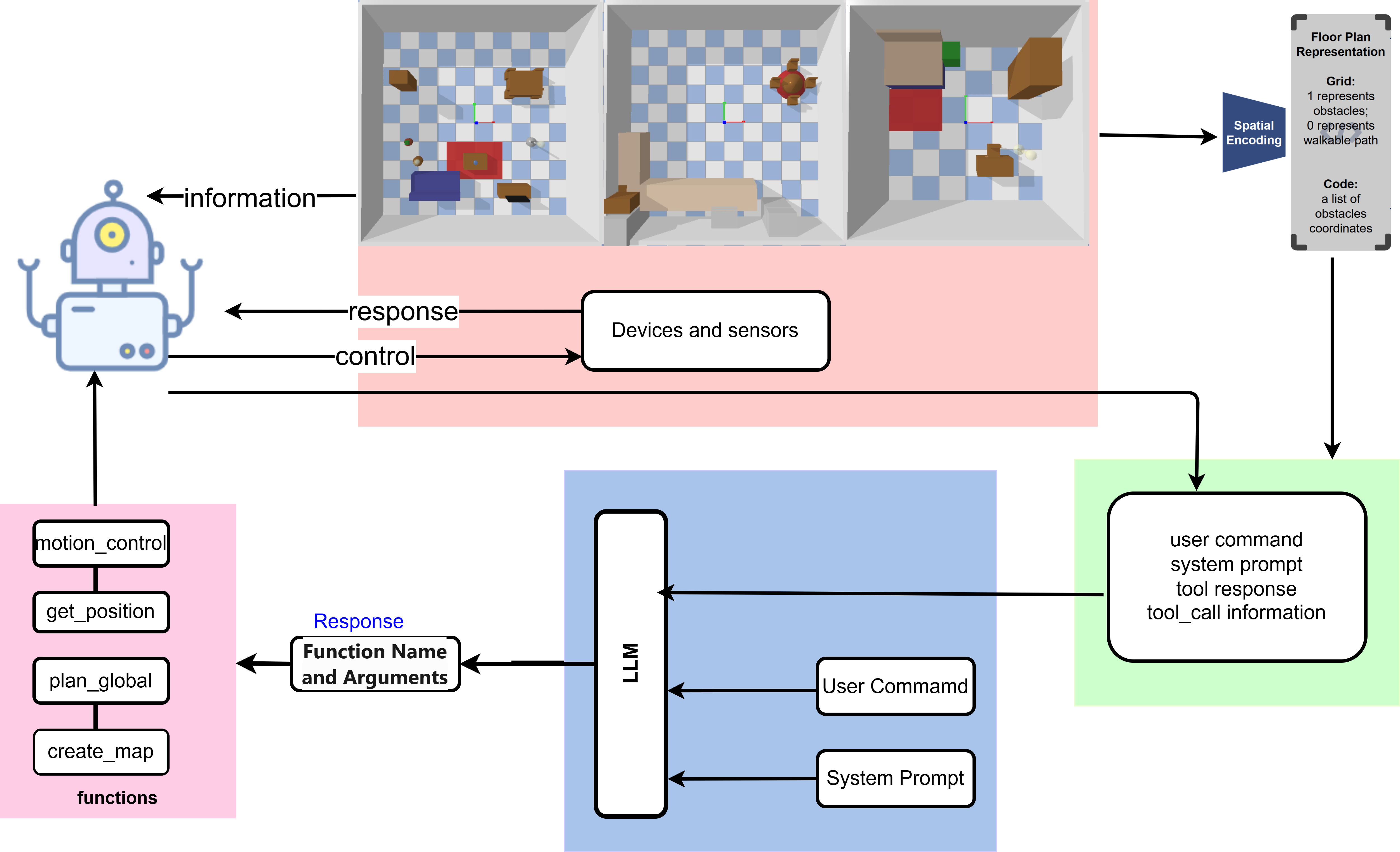}
  \caption{Framework }
  \label{fig:framework}
\end{figure}

As illustrated in Figure~\ref{fig:framework}, our framework employs a recursive execution loop where user commands are concatenated with system prompts and processed by the LLM, which parses the input to identify appropriate functions and parameters. The robotic system executes these functions and feeds the results alongside contextual history back into the LLM for iterative analysis until navigation task completion.
Experimental validation was conducted in PyBullet---a Python-based interface to the Bullet physics engine specifically designed for robotics simulation, game development, and reinforcement learning research. This implementation achieves two critical objectives: (1) systematically modularizing traditional navigation functionalities (e.g., path planning, obstacle avoidance) into atomic operations with standardized interfaces, and (2) designing a hierarchical prompt architecture and interaction protocol that resolves spatial referential ambiguities through context-aware reasoning. The modular function library enables environment-specific adaptability through flexible component updates, while the multi-stage prompt engineering enhances the LLM's capacity to interpret environmental semantics and user intent, thereby bridging linguistic instructions with robotic actions in physics-accurate simulations.

\setlength{\parindent}{0.51cm} 
\subsection{Functions}
\subsubsection{location} 
    Simulation environments provide straightforward access to robotic positional data through built-in functions. For instance, the \texttt{get\_husky\_position} function retrieves an object’s (e.g., a robot) world frame coordinates (\texttt{pos}) and orientation quaternion (\texttt{orn}), offering precise spatial state information without requiring external sensors. This functionality leverages the simulator’s inherent capacity to track ground-truth poses, making it invaluable for tasks like motion planning validation, localization algorithm testing, and real-time performance monitoring in controlled virtual settings.
\subsubsection{Map}
    During simulation environment configuration, the system generates extensive raw data, which undergoes cleaning and structuring to derive comprehensive spatial map representations. This processed data is then exported to a JSON file, serving as a foundational resource for downstream path planning operations. A pivotal step involves the \texttt{world\_to\_grid(pos, resolution)} function, which converts world-frame coordinates (in meters) to grid map indices (in pixels/grid units). During spatial data processing, grid cells containing objects are flagged as obstacles (assigned a value of 1), with an inflation radius applied to surrounding cells to ensure safety margins, while unoccupied areas remain marked as 0. This workflow enables efficient environment modeling for autonomous navigation and collision avoidance.

Among various alternatives, the most commonly used and intuitive representation of space is in the form of grids \cite{ref2,ref5,ref16}. In this representation, the environment is discretized into a grid structure where each cell corresponds to a specific location, allowing for clear and precise definitions of free spaces, obstacles, and agent positions.

Alternatively, space can be represented using code-based descriptions \cite{ref19}, which can be more interpretable for LLMs. This approach is both compact and flexible, enabling precise definitions of the environment through code. For instance, variables can be defined to specify the start and goal locations, while logic can be applied to place obstacles on the grid, shaping the environment accordingly. Intuitively, code provides a clear and concise way to define the task setting, making it a powerful alternative to traditional grid-based representations.

By using text-based representations, we bridge the gap between spatial reasoning and natural language processing, enabling LLMs to leverage their reasoning capabilities in a domain where they have proven effectiveness. This text-based approach sets ourselves apart from images as input, which may introduce unnecessary or redundant information such as textures, colors, or irrelevant details, while enabling the language capabilities of LLMs, which excel at processing and reasoning with text. To formalize, we define a grid-based environment representation:
$G = \{g_{i,j} \mid g_{i,j} \in \{0, 1\}\}$
where \(1\) indicates an obstacle, \(0\) denotes free space, and the element represents the cell at row \(i\) and column \(j\) in a 2D grid.

We also define a code representation as a list of obstacle coordinates:
\[
C = \texttt{obstacles.append}((i_1, j_1),\, \ldots,\, (i_n, j_n))
\]
where each \((i, j)\) denotes the location of an obstacle.
\subsubsection{Planning}
    Building upon the acquired grid map data, this implementation centers on two key functional components. The path planning system employs an 8-directional A* search algorithm, operationalized through two core functions: the \texttt{get\_neighbors} function identifies traversable neighboring nodes from the current grid position, while the \texttt{a\_star} function implements the core algorithmic logic with integrated heuristic cost calculation and priority queue management. This dual-module architecture ensures efficient exploration of grid-based environments while preserving the theoretical optimality guarantees of the A* framework.

The \texttt{get\_neighbors} function identifies traversable neighboring nodes from the current grid position through the following workflow:
\begin{enumerate}[label=\arabic*., itemsep=0.3ex, parsep=0.3ex]
  \item 8-directional coordinate generation: Calculate potential positions using directional offsets (e.g., $\pm 1$ grid steps horizontally/vertically/diagonally).
  \item Boundary and accessibility validation: Discard positions exceeding grid boundaries or occupying obstructed cells (grid value $\neq 0$).
  \item Collision-aware diagonal filtering: Conditionally exclude diagonal moves (e.g., $\nearrow$) if adjacent cardinal-direction cells (e.g., right or up) are obstructed (value = 1), thereby preventing invalid ``corner-cutting'' through obstacles.
  \item Traversable node compilation: Return a validated list of coordinates meeting all kinematic and environmental constraints.
\end{enumerate}

The \texttt{a\_star} function implements the core logic of the A* algorithm through the following workflow: It initializes a priority queue while recording the starting point with zero cost, and establishes a backtracking dictionary for path reconstruction. The algorithm then iteratively extracts the highest-priority node (with the smallest sum of accumulated cost and heuristic value) from the queue. If the target node is reached, the search terminates; otherwise, it examines all valid neighboring nodes. Movement costs are calculated using $\sqrt{2}$ for diagonal moves and \(1\) for straight moves. When a more optimal path is detected, the algorithm updates the cost records and requeues the node with recalculated priority. Finally, it reconstructs the path using the backtracking dictionary and returns the validated result after confirming path accessibility from the starting point.

Key features of this implementation include eight-directional movement support, dynamic obstacle detection, path cost optimization, and heuristic-guided search acceleration. The algorithm's robustness stems from two critical design elements: specialized handling of diagonal movement costs and rigorous path validity verification, which collectively ensure both computational efficiency and solution accuracy.

\begin{algorithm}[tb]  
\caption{A* Path Planning (8-directional)}
\label{alg:astar}
\small  
\begin{algorithmic}[1]
\Procedure{A*}{start, goal, grid}
\State \texttt{open\_list} $\gets$ priority queue
\State \texttt{g\_cost} $\gets$ \{\texttt{start}:0\}, \texttt{parent} $\gets$ \{\}
\State enqueue \texttt{start} with $f = 0 + h(\texttt{start}, \texttt{goal})$

\While{open list not empty}
    \State $u \gets$ dequeue min $f$
    \If{$u = \texttt{goal}$}
        \State \Return \texttt{reconstruct\_path(parent)}
    \EndIf
    
    \ForAll{$v \in \text{Get\_Neighbors}(u, \texttt{grid})$}
        \State $c(u \to v) = 
        \begin{cases} 
            \sqrt{2} & \text{diagonal} \\
            1 & \text{orthogonal}
        \end{cases}$
        \State $\texttt{g\_tentative} = \texttt{g\_cost}[u] + c$
        
        \If{$\texttt{g\_tentative} < \texttt{g\_cost}.get(v, \infty)$}
            \State update \texttt{g\_cost}[v] $\gets$ tentative
            \State $f(v) = \texttt{g\_cost}[v] + h(v, \texttt{goal})$
            \State enqueue/update $v$ in \texttt{open\_list}
            \State $\texttt{parent}[v] \gets u$
        \EndIf
    \EndFor
\EndWhile
\State \Return \texttt{empty path}
\EndProcedure
\end{algorithmic}
\end{algorithm}

\subsubsection{Motion\_control}
    This study presents an integrated differential drive control architecture that combines path planning with dynamic adjustment, implementing waypoint navigation through PID control. The system employs a hierarchical control strategy: an upper layer generates global paths using an enhanced A* algorithm with proactive waypoint optimization, while a lower layer executes trajectory tracking through an adaptive PID controller. The core workflow operates as follows:
\begin{itemize}[leftmargin=*,itemsep=0pt]
    \item[] \textbf{(1) Path Preprocessing} \\
    The original path point sequence generated by the A\textsuperscript{*} algorithm undergoes simplification and optimization through:
    \begin{itemize}[leftmargin=*,label=\textbullet]
        \item \textbf{5-step look-ahead mechanism}: Collision detection validates straight-line traversability across a predefined horizon
        \item \textbf{Jump connection strategy}: Reduces waypoints by $\sim$45\% (Formula~X), significantly lowering control complexity
    \end{itemize}
\end{itemize}

\begin{itemize}[leftmargin=*,itemsep=0pt,topsep=0pt]
    \item[] \textbf{(2) PID Error Feedback Control}
    \begin{itemize}[left=5pt,noitemsep,topsep=-3pt]
        \item[] \textbf{Target-point tracking error model:}
        \vspace{-0.8ex}
        \begin{equation*}
            \begin{cases}
                e_d = \| \mathbf{p}_k - \mathbf{q}_t \|_2 & \text{(distance error)} \\
                e_\theta = \arctan(\frac{\Delta y}{\Delta x}) - \theta_t & \text{(heading error)}
            \end{cases}
        \end{equation*}
        \begin{itemize}[left=10pt,noitemsep,topsep=-5pt]
            \item[] \textbf{Distance error $e_d$}: Euclidean distance between $\mathbf{p}_k$ and $\mathbf{q}_t$
            \item[] \textbf{Heading error $e_\theta$}: Angular difference between $\arctan(\frac{\Delta y}{\Delta x})$ and $\theta_t$
        \end{itemize}
        
        \item[] \textbf{Discrete PID controller:} Steering command
        \vspace{-1.2ex}
        \begin{equation*}
            u_\theta = K_p e_\theta + K_i \sum_{i=0}^t e_\theta(i)\Delta t + K_d \frac{e_\theta(t) - e_\theta(t-1)}{\Delta t}
        \end{equation*}
        \begin{itemize}[left=10pt,noitemsep,topsep=-5pt]
            \item[] \textbf{Parameter tuning}: PSO optimized ($K_p=3.2,\ K_i=0.1,\ K_d=0.3$)
            \item[] \textbf{Output constraint}: $\delta \in [-\frac{\pi}{5}, \frac{\pi}{5}]$
        \end{itemize}
    \end{itemize}
\end{itemize}

\begin{itemize}[leftmargin=*,noitemsep,topsep=0pt]
    \item[] \textbf{(3) Dynamic Velocity Modulation}
    \begin{itemize}[left=5pt,noitemsep,topsep=-3pt]
        \item[] \textbf{Speed-steering coupling strategy:}\vspace{-0.5ex}
        $$
        v_{\mathrm{base}} = v_0 \cdot \underbrace{\min\left(\frac{e_d}{1.2},1\right)}_{\text{Distance Decay}} \cdot \underbrace{\left(1-\frac{|e_\theta|}{\pi/3}\right)}_{\text{Braking}} \cdot 1.2
        $$
        \item[] \textbf{Reverse mode activation:}\vspace{-0.8ex}
        $$
        |e_\theta| > \frac{\pi}{2} \Rightarrow v_{\mathrm{base}} \leftarrow -0.6v_{\mathrm{base}},\ \delta \leftarrow -\delta
        $$
    \end{itemize}
\end{itemize}
The velocity modulation model dynamically adjusts the baseline velocity through coupling path-tracking errors, where the distance decay factor limits the maximum operational velocity and the heading error compensation term implements safety braking. When heading deviation exceeds 90°, the reverse motion mode is triggered to ensure obstacle avoidance capability via velocity inversion and steering angle mirroring.
\begin{itemize}[leftmargin=*,itemsep=0pt,topsep=0pt]
    \item[] \textbf{(4) Differential Steering Implementation}
    \begin{itemize}[left=5pt,noitemsep=0pt,topsep=-3pt]
        \item[] \textbf{Smooth control via hyperbolic tangent:}
        \vspace{-1.2ex}
        $$
        \begin{cases}
            V_L = V_{\mathrm{base}}(1 - \tanh(\delta)) \\
            V_R = V_{\mathrm{base}}(1 + \tanh(\delta))
        \end{cases}
        $$      
    \end{itemize}
\end{itemize}
Performance improvement: Reduces lateral acceleration fluctuations compared to conventional linear mapping.
\begin{itemize}[leftmargin=*,itemsep=0pt,topsep=0pt]
    \item[] \textbf{(5) Waypoint Transition Mechanism}
    \begin{itemize}[left=5pt,noitemsep=0pt,topsep=-3pt]
        \item[] \textbf{Multi-constraint arrival criterion:}
        \vspace{-1.2ex}
        $$
e_d < 0.4\,\mathrm{m} \ \land \ |e_{\theta}| < \frac{\pi}{3} \ \land \ \|v\| < 1\,\mathrm{m/s}
        $$      
    \end{itemize}
\end{itemize}
Functional enhancement: Eliminates oscillatory switching behavior with improved success rate.

\subsection{system prompt}
This section focuses on system prompt design, which follows a framework structurally analogous to the MCP protocol. The discussion is organized into two principal components: (1) the constituent elements and functional specifications of system prompt content, and (2) the interaction workflow with large language models (LLMs).
\subsubsection{Composition and Specifications of System Prompt Content}
    
\begin{enumerate}[
  label=\arabic*., 
  leftmargin=*,        
  itemsep=0pt,         
  parsep=0pt,          
  topsep=0pt,          
  partopsep=14pt,       
  labelsep=0.5em,      
  before=\vspace{-0.5\baselineskip}, 
  after=\vspace{-0.5\baselineskip}   
]

  \item \textbf{Context and Role Definition}\\
  The system establishes an expert role specialized in robotic navigation, explicitly defining operational boundaries across core competencies including localization, mapping, path planning, and motion control. This role operates under stepwise reasoning principles to achieve autonomous decision-making through systematic tool invocation.
  \item \textbf{Tool Utilization Logic}\\
  A single-step iterative mechanism governs tool usage, requiring sequential execution of individual tools with mandatory feedback verification. This creates an ``execution-feedback-decision'' closed-loop control structure to ensure task progression reliability.
  
  \item \textbf{Tool Invocation Syntax}\\
  An XML-based standardized template enforces hierarchical encapsulation of tool names and parameters within dedicated tags. This syntax ensures command parsability and cross-system compatibility through rigorous format validation protocols.
  
  \item \textbf{Tool Operation Protocol}\\
  A four-phase decision workflow is mandated: environmental analysis, tool selection, parameter validation, and result response. Reasoning processes must be explicitly documented via XML tags, with built-in mechanisms for operation interruption during parameter deficiencies and multi-step dependency resolution.
  
  \item \textbf{Core Toolset Configuration}\\
  Five functional primitives are provisioned with parameter constraints and default rules: environment modeling (\texttt{create\_grid\_map}), global path planning (\texttt{plan\_global\_path}), motion control (\texttt{motion\_control}), state monitoring (\texttt{get\_husky\_position}), and environmental perception (\texttt{get\_living\_room\_info}). The architecture supports modular tool replacement or extension through standardized interface definitions.
  
  \item \textbf{Task Execution Framework}\\
  A four-layer hierarchical architecture is implemented:
  \begin{itemize}
    \item Task Decomposition Engine generates prioritized subgoal sequences through logical stratification.
    \item Tool-Driven Execution Controller enforces feedforward control to align tool invocation sequences with dataflow requirements.
    \item Pre-Execution Validation Module integrates contextual analysis, tool suitability evaluation, and parameter completeness verification.
    \item Closed-Loop Output Interface mandates standardized solution submission with full execution traceability.
  \end{itemize}
  
  \item \textbf{Physical Constraints}\\
  Non-negotiable boundary conditions are enforced: spatial resolution (0.05m), maximum velocity (\(\leq 40\)m/s), and robot ID (1). These constraints serve as immutable parameters during tool validation.

  \item \textbf{Reference Workflow}\\
  A canonical navigation pipeline is demonstrated: environmental sensing \(\rightarrow\) grid mapping (\texttt{create\_grid\_map}) \(\rightarrow\) A* path planning (\texttt{plan\_global\_path}) \(\rightarrow\) motion execution (\texttt{motion\_control}) \(\rightarrow\) pose monitoring (\texttt{get\_husky\_position}). The workflow explicitly defines tool invocation sequences and inter-tool dataflow patterns.
\end{enumerate}
The proposed architecture employs XML-structured instructions to encapsulate navigation toolchain operations. Through \texttt{<tool\_chain>} tags for functional composition, \texttt{<feedback\_loop>} for error tracing, and \texttt{<dynamic\_adjustment>} for policy iteration, it establishes a reusable framework with standardized interfaces. This design reduces system migration complexity while maintaining extensibility for tool replacement and multi-scenario adaptation.

\subsubsection{Interaction Workflow of Large Language Models}
\begin{figure}[ht]
  \centering
  \includegraphics[width=\linewidth]{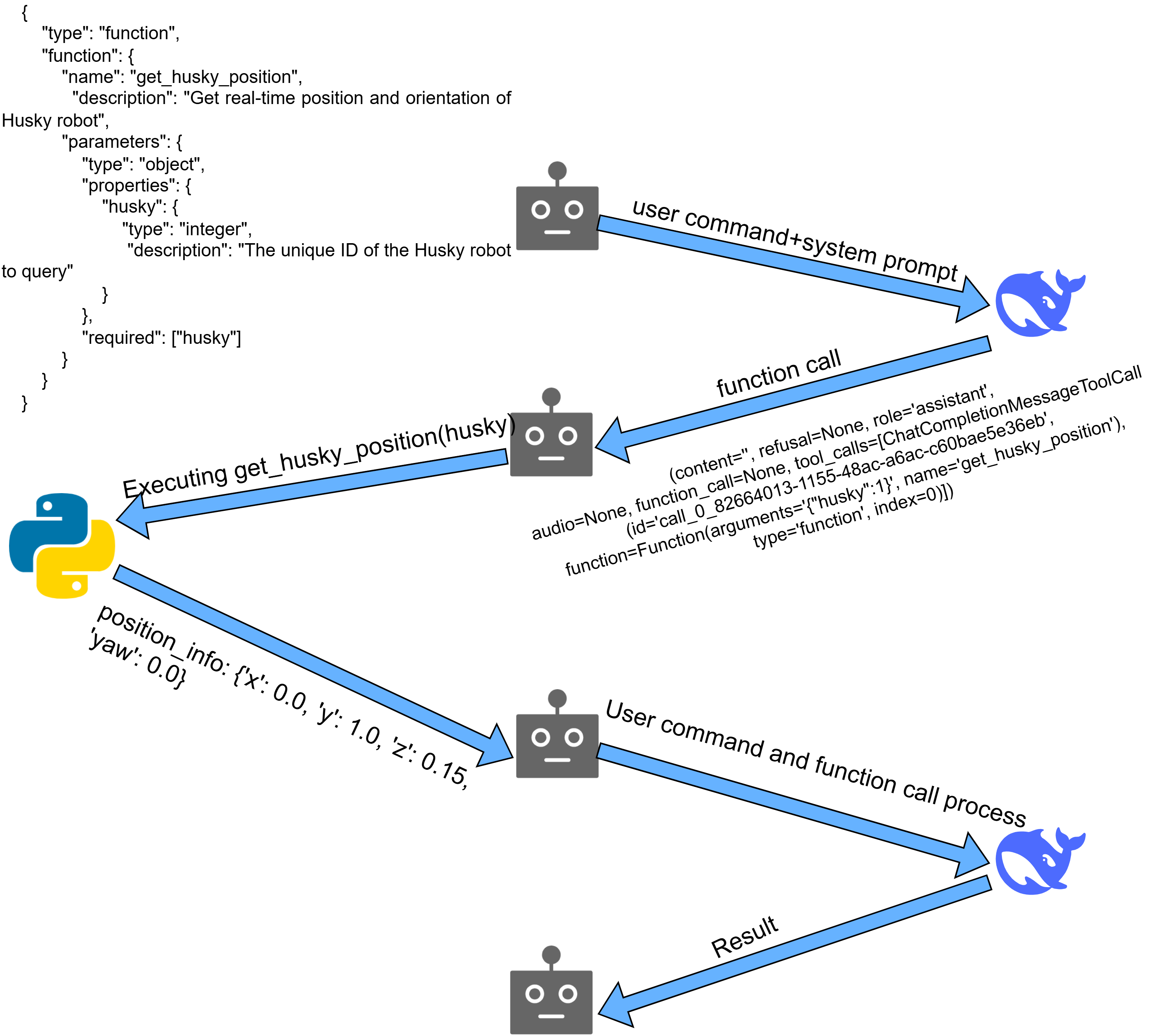}
  \caption{Interaction workflow diagram (Figure 2).}
  \label{fig:workflow}
\end{figure}

    As illustrated in Figure~\ref{fig:workflow}, the diagram demonstrates a single interaction cycle between a Large Language Model (LLM) and a robotic system, where multiple iterative exchanges are typically required to accomplish complex tasks. The architecture comprises three core components: the LLM (right), the robotic interface (center), and Python-implemented functional modules (left).

The workflow initiates with a user submitting natural language commands. These commands are subsequently combined with predefined system prompts and transmitted to the LLM. Through semantic parsing of both the user input and system directives, the LLM generates structured function calls containing operational parameters, which are then dispatched to the robotic system. Upon receiving these instructions, the robotic subsystem executes the corresponding Python functions through its application programming interface (API). Following task execution, the system aggregates operational feedback, historical interaction data, and the original user command into a consolidated input package for subsequent LLM processing. The LLM continuously evaluates task completion status through this cyclical interaction pattern, persisting in function call generation until all operational objectives are fulfilled.

\section{Results \& Discussion}
\subsection{Implementation Details }
    Our experimental environment is constructed using PyBullet simulation platform, incorporating three distinct domestic scenarios: living room, kitchen, and bedroom. The large language model (LLM) infrastructure integrates API endpoints from DeepSeek-v3 and OpenAI-4o mini, complemented by a locally deployed Phi-4-14B model. Our comprehensive benchmarking analysis compared three state-of-the-art language models: Phi-4-14B, Llama-3-8B, and Qwen-2.5-14B. The Phi-4-14B architecture demonstrated superior performance across multiple test suites, achieving comparable results to the significantly larger Llama-3-70B-Instruct model. Notably, Phi-4 exhibited exceptional mathematical reasoning capabilities, outperforming both GPT-4o and Gemini Pro 1.5 by substantial margins on the GPQA (Generalized Problem Question Answering) and MATH (Mathematical Aptitude Test for Heuristics) benchmarks. This empirical evidence substantiates our selection of Phi-4 as the optimal model for deployment.The methodological framework encompasses two navigation paradigms: conventional navigation our proposed LLM agent-driven navigation.

The dataset configuration for living room scenario comprises: 1) Starting positions: 10 uniformly distributed representative locations; 2) Target positions: 60 randomly selected destinations within the spatial boundaries. Each navigation method undergoes 600 experimental trials. Notably, approximately 18.3\% of target positions exhibit accessibility constraints due to static obstacles (e.g., furniture such as tables, sofas, and cabinets), consequently reducing task completion rates.

Implementation challenges primarily stem from two aspects: 1) Physical constraints of the robotic agent in simulation, including non-negligible body dimensions and control system tolerances; 2) Suboptimal performance in edge cases despite parameter optimization in motion control functions. These factors collectively contribute to navigation failures even when employing theoretically sound path planning algorithms.

\subsection{Metrics}
    Trajectory Length (TL), denoting the average distance traveled by the agent; Navigation Error (NE), representing the mean distance from the agent’s final location to the destination; Success Rate (SR), indicating the proportion of navigation episodes where the agent successfully reaches the target location within a 0.5m margin of error; Instruction Understanding Success (SU), measuring the proportion of verbal commands correctly parsed and translated into executable actions during human-AI interaction phases; Path Length (PL), denoting the ratio of the user’s actual traveled path length to the theoretical shortest path, reflecting the path optimization capability of the navigation system; and Success Rate weighted by the normalized inverse of Path Length (SPL), which is a more nuanced measure that balances navigation precision and efficiency by adjusting the success rate based on the ratio of the optimal path length to the agent’s predicted path length.

\subsection{Discussion}
\begin{table}[!ht]
\centering
\fontfamily{ptm}\selectfont 
\small 
\caption{Performance comparison under different scenarios}
\label{tab:performance}
\setlength{\tabcolsep}{3.5pt} 
\hspace*{-0.05\columnwidth} 
\begin{tabular}{@{}llccccccc@{}}
\toprule
\multicolumn{1}{c}{\textbf{Scene}} & \textbf{Method} & \textbf{TL} & \textbf{NE} & \textbf{SR} & \textbf{PL} & \textbf{SPL} & \textbf{SU} \\ 
\midrule
\multirow{2}{*}{Living room} 
& Navigation  & 5.19 & 0.484 & 78.5  & 1.09  & 0.926 & --    \\
& Nav(ours)   & 5.23 & 0.461 & 78.1  & 1.06  & 0.921 & 0.95  \\
\cmidrule(r){1-8}
\multirow{2}{*}{Kitchen}
& Navigation  & 4.13 & 0.477 & 80.3  & 1.15  & 0.925 & --    \\
& Nav(ours)   & 4.12 & 0.465 & 80.1  & 1.14  & 0.925 & 0.96  \\
\cmidrule(r){1-8}
\multirow{2}{*}{Bedroom}
& Navigation  & 3.05 & 0.479 & 81.2  & 1.25  & 0.873 & --    \\
& Nav(ours)   & 3.13 & 0.484 & 80.3  & 1.27  & 0.871 & 0.96  \\
\bottomrule
\end{tabular}
\end{table}
\fontfamily{\familydefault}\selectfont 
\noindent Tab.1 presents the evaluation metrics of different navigation methods across three scenarios. Our method in Table I utilizes the API based on the DeepSeek-v3 model. The experimental scenarios include the living room, kitchen, and bedroom, with comparative methods comprising conventional navigation and our approach derived from functional decomposition of conventional navigation. The units for TL and NE in Table I are meters, SR is expressed as a percentage, while PL, SPL, and SU are dimensionless ratios.
In the spatially larger living room environment, the trajectory length (TL) measurements in the dataset show values averaging 1 meter greater than those in the kitchen and 2 meters greater than in the bedroom, aligning with spatial reality. Navigation errors (NE) across all three environments remain within 0.5 meters. The success rates (SR) of our method closely align with conventional navigation, as expected given our functional decomposition framework building upon traditional navigation. Path length (PL) and success weighted by path length (SPL) metrics demonstrate comparable performance across methods.
Notably, our method exhibits overwhelming superiority in Instruction Understanding Success (SU), - a critical capability reflecting system generalization. This advantage stems primarily from the extensive knowledge inherent in large language models (LLMs), which fundamentally motivates their integration as navigation controllers.

\begin{table}[!ht]
\centering
\fontfamily{ptm}\selectfont\small
\caption{Comparative Analysis Across Multiple Large Language Model Architectures}
\label{tab:comparison}
\setlength{\tabcolsep}{4.5pt}
\hspace*{0\columnwidth} 
\begin{tabular}{@{}llcccccc@{}}
\toprule
\multicolumn{1}{@{}l}{\textbf{Scene}} & \textbf{Method} & \textbf{TL} & \textbf{NE} & \textbf{SR} & \textbf{PL} & \textbf{SPL} & \textbf{SU} \\ 
\midrule
\multirow{3}{*}{Living room} 
& DeepSeek  & 5.23 & 0.461 & 78.1  & 1.06  & 0.921 & 0.95 \\
& OpenAi    & 5.34 & 0.491 & 75.5  & 1.18  & 0.912 & 0.86 \\
& Phi-4     & 5.37 & 0.488 & 60.5  & 1.24  & 0.804 & 0.66 \\
\cmidrule(r){1-8}
\multirow{3}{*}{Kitchen}
& DeepSeek  & 4.12 & 0.468 & 80.1  & 1.14  & 0.925 & 0.96 \\
& OpenAi    & 4.11 & 0.467 & 76    & 1.12  & 0.93  & 0.87 \\
& Phi-4     & 4.25 & 0.477 & 61    & 1.29  & 0.812 & 0.67 \\
\cmidrule(r){1-8}
\multirow{3}{*}{Bedroom}
& DeepSeek  & 3.13 & 0.484 & 80.3  & 1.27  & 0.871 & 0.96 \\
& OpenAi    & 3.08 & 0.465 & 78    & 1.2   & 0.89  & 0.88 \\
& Phi-4     & 3.25 & 0.487 & 62    & 1.3   & 0.798 & 0.65 \\
\bottomrule
\end{tabular}
\end{table}
\fontfamily{\familydefault}\selectfont

\noindent Tab.2 illustrates the impact of different models on our method. The experimental scenarios remain consistent with the living room, kitchen, and bedroom environments, employing three comparative models: DeepSeek-v3, OpenAI-4o Mini, and Phi-4 (14B version). Metrics including TL, NE, PL, and SPL demonstrate comparable performance across all models. Regarding success rate (SR), DeepSeek-v3 marginally outperforms OpenAI-4o Mini while exhibiting nearly 20\% superiority over Phi-4. Instruction Understanding Success (SU), DeepSeek-v3 achieves a 10\% advantage over OpenAI-4o Mini and a 30\% improvement compared to Phi-4.

This performance discrepancy primarily correlates with the parametric scale of the LLMs. DeepSeek-v3, though parametrically larger than OpenAI-4o Mini, shows only <10\% performance gaps in SR and SU due to their comparable model capabilities. In contrast, OpenAI-4o Mini exceeds Phi-4 by multiple orders of magnitude in parameter count, resulting in >15\% performance advantages in both SR and SU. Notably, Phi-4 demonstrates unique deployability advantages for local implementation scenarios – a logistical benefit not shared by larger parametric models.

\section{Conclusions}
In this work, we investigate the potential of large language models (LLMs) in embodied navigation tasks. We propose a navigation framework leveraging LLM function calling mechanisms, which achieves open-world semantic comprehension capabilities while enabling modular extensibility of tools. The reusable system prompt templates empower robotic agents to effectively address navigation challenges in unfamiliar environments.

Nevertheless, three critical constraints persist: (1) the granularity and implementation of functional decomposition, (2) inherent limitations in LLMs' cognitive capacities, and (3) sensitivity to interaction protocols and prompt engineering. While our methodology demonstrates LLMs' substantial potential in navigation systems, these factors collectively bound their operational efficacy.

Our findings substantiate LLMs as foundational components for next-generation intelligent agents, charting a pivotal direction for embodied AI systems.

\bibliographystyle{plain} 
\bibliography{references}    

\begin{thebibliography}{10}

\bibitem{ref2}
Mohamed Aghzal, Erion Plaku, and Ziyu Yao.
\newblock Look further ahead: Testing the limits of gpt-4 in path planning.
\newblock In {\em 2024 IEEE 20th International Conference on Automation Science and Engineering (CASE)}, pages 1020--1027. IEEE, 2024.

\bibitem{ref19}
Mohamed Aghzal, Erion Plaku, and Ziyu Yao.
\newblock Look further ahead: Testing the limits of gpt-4 in path planning.
\newblock In {\em 2024 IEEE 20th International Conference on Automation Science and Engineering (CASE)}, pages 1020--1027. IEEE, 2024.

\bibitem{ref20}
M.~Ahn, A.~Brohan, N.~Brown, Y.~Chebotar, O.~Cortes, B.~David, C.~Finn, C.~Fu, K.~Gopalakrishnan, K.~Hausman, A.~Herzog, D.~Ho, J.~Hsu, J.~Ibarz, B.~Ichter, A.~Irpan, E.~Jang, R.~J. Ruano, K.~Jeffrey, S.~Jesmonth, N.~Joshi, R.~Julian, D.~Kalashnikov, Y.~Kuang, K.-H. Lee, S.~Levine, Y.~Lu, L.~Luu, C.~Parada, P.~Pastor, J.~Quiambao, K.~Rao, J.~Rettinghouse, D.~Reyes, P.~Sermanet, N.~Sievers, C.~Tan, A.~Toshev, V.~Vanhoucke, F.~Xia, T.~Xiao, P.~Xu, S.~Xu, M.~Yan, and A.~Zeng.
\newblock Do as i can and not as i say: Grounding language in robotic affordances.
\newblock arXiv preprint arXiv:2204.01691, 2022.

\bibitem{ref4}
S.~Bubeck, V.~Chandrasekaran, R.~Eldan, J.~Gehrke, E.~Horvitz, E.~Kamar, P.~Lee, Y.~T. Lee, Y.~Li, S.~Lundberg, et~al.
\newblock Sparks of artificial general intelligence: Early experiments with gpt-4.
\newblock arXiv preprint arXiv:2303.12712, 2023.

\bibitem{ref9}
A.~Chowdhery, S.~Narang, J.~Devlin, M.~Bosma, G.~Mishra, A.~Roberts, P.~Barham, H.~W. Chung, C.~Sutton, S.~Gehrmann, et~al.
\newblock Palm: Scaling language modeling with pathways.
\newblock arXiv preprint arXiv:2204.02311, 2022.

\bibitem{ref13}
D.~Driess, F.~Xia, M.~S. Sajjadi, C.~Lynch, A.~Chowdhery, B.~Ichter, A.~Wahid, J.~Tompson, Q.~Vuong, T.~Yu, et~al.
\newblock Palm-e: An embodied multimodal language model.
\newblock arXiv preprint arXiv:2303.03378, 2023.

\bibitem{ref18}
S.~Y. Gadre, M.~Wortsman, G.~Ilharco, L.~Schmidt, and S.~Song.
\newblock Clip on wheels: Open-vocabulary models are (almost) zero-shot object navigators.
\newblock arXiv, 2022.

\bibitem{ref12}
C.~Huang, O.~Mees, A.~Zeng, and W.~Burgard.
\newblock Visual language maps for robot navigation.
\newblock arXiv preprint arXiv:2210.05714, 2022.

\bibitem{ref17}
A.~Majumdar, G.~Aggarwal, B.~Devnani, J.~Hoffman, and D.~Batra.
\newblock Zson: Zero-shot object-goal navigation using multimodal goal embeddings.
\newblock arXiv preprint arXiv:2206.12403, 2022.

\bibitem{ref11}
D.~Shah, B.~Osinski, S.~Levine, et~al.
\newblock Llm-nav: Robotic navigation with large pre-trained models of language, vision, and action.
\newblock In {\em Conference on Robot Learning}, pages 492--504. PMLR, 2023.

\bibitem{ref10}
Y.~Tao, X.~Liu, I.~Spasojevic, S.~Agarwal, and V.~Kumar.
\newblock 3d active metric-semantic slam.
\newblock {\em IEEE Robotics and Automation Letters}, 9(3):2989--2996, 2024.

\bibitem{ref7}
S.~Vemprala, R.~Bonatti, A.~Bucker, and A.~Kapoor.
\newblock Chatgpt for robotics: Design principles and model abilities, 2023.

\bibitem{ref15}
J.~Wei, Y.~Tay, R.~Bommasani, C.~Raffel, B.~Zoph, S.~Borgeaud, D.~Yogatama, M.~Bosma, D.~Zhou, D.~Metzler, et~al.
\newblock Emergent abilities of large language models.
\newblock arXiv preprint arXiv:2206.07682, 2022.

\bibitem{ref8}
Congcong Wen, Yisiyuan Huang, Hao Huang, Yanjia Huang, Shuaihang Yuan, Yu~Hao, Hui Lin, {Yu-Shen} Liu, and Yi~Fang.
\newblock Zero-shot object navigation with vision-language models reasoning.
\newblock {\em arXiv preprint arXiv:2410.18570}, 2024.
\newblock Available at: \url{https://arxiv.org/abs/2410.18570v1}.

\bibitem{ref6}
J.~Wu, X.~Li, S.~Xu, H.~Yuan, H.~Ding, Y.~Yang, X.~Li, J.~Zhang, Y.~Tong, X.~Jiang, B.~Ghanem, and D.~Tao.
\newblock Towards open vocabulary learning: A survey.
\newblock {\em IEEE Transactions on Pattern Analysis and Machine Intelligence}, 46(7):5092--5113, 2024.

\bibitem{ref5}
Wenshan Wu, Shaoguang Mao, Yadong Zhang, Yan Xia, Li~Dong, Lei Cui, and Furu Wei.
\newblock Mind’s eye of llms: Visualization-of-thought elicits spatial reasoning in large language models.
\newblock In {\em The Thirty-eighth Annual Conference on Neural Information Processing Systems}, 2024.

\bibitem{ref16}
Jihan Yang, Shusheng Yang, Anjali~W. Gupta, Rilyn Han, Li~Fei-Fei, and Saining Xie.
\newblock Thinking in space: How multimodal large language models see, remember, and recall spaces.
\newblock 2024.

\bibitem{ref1}
Hang Yin, Xiuwei Xu, Zhenyu Wu, Jie Zhou, and Jiwen Lu.
\newblock Sg-nav: Online 3d scene graph prompting for llm-based zero-shot object navigation.
\newblock In {\em Advances in Neural Information Processing Systems (NeurIPS)}, 2024.

\bibitem{ref3}
Hang Yin, Xiuwei Xu, Linqing Zhao, Ziwei Wang, Jie Zhou, and Jiwen Lu.
\newblock Unigoal: Towards universal zero-shot goal-oriented navigation.
\newblock {\em arXiv preprint arXiv:2503.10630v3}, 2025.

\bibitem{ref14}
K.~Zhou, K.~Zheng, C.~Pryor, Y.~Shen, H.~Jin, L.~Getoor, and X.~E. Wang.
\newblock Esc: Exploration with soft commonsense constraints for zero-shot object navigation.
\newblock arXiv preprint arXiv:2301.13166, 2023.

\end{thebibliography}
\end{document}